\pdfoutput=1
\documentclass[11pt]{article}
\usepackage[final]{acl}
\usepackage{times}
\usepackage{latexsym}
\usepackage[T1]{fontenc}
\usepackage[utf8]{inputenc}
\usepackage{microtype}
\usepackage{inconsolata}
\usepackage{graphicx} 
\usepackage{amsmath}
\usepackage{booktabs}
\usepackage{siunitx}
\usepackage{changes}
\usepackage{multirow}

\title{
Context is Important in Depressive Language: A Study of the Interaction Between the Sentiments and Linguistic Markers in Reddit Discussions}
\author{Neha Sharma \and Kairit Sirts \\
  Institute of Computer Science, University of Tartu, Estonia\\
  \texttt{neha.sharma@ut.ee}, \texttt{sirts@ut.ee} \\}

\begin{document}
\maketitle
\begin{abstract}
Research exploring linguistic markers in individuals with depression has demonstrated that language usage can serve as an indicator of mental health. This study investigates the impact of discussion topic as context on linguistic markers and emotional expression in depression, using a Reddit dataset to explore interaction effects. Contrary to common findings, our sentiment analysis revealed a broader range of emotional intensity in depressed individuals, with both higher negative and positive sentiments than controls. This pattern was driven by posts containing no emotion words, revealing the limitations of the lexicon based approaches in capturing the full emotional context. We observed several interesting results demonstrating the importance of contextual analyses. For instance, the use of 1st person singular pronouns and words related to anger and sadness correlated with increased positive sentiments, whereas a higher rate of present-focused words was associated with more negative sentiments. Our findings highlight the importance of discussion contexts while interpreting the language used in depression, revealing that the emotional intensity and meaning of linguistic markers can vary based on the topic of discussion.
\end{abstract}

\section{Introduction}
Research on linguistic markers of depression aiming to identify people at risk \citep{demunmun2013paper,yates2017depression,copper2018,chancellor2020methods} have found that individuals with depression often use more first-person pronouns and negative emotion words and fewer positive emotion words than healthy controls \citep{trifu2017linguistic,savekar2023structural,rude2004language,chung2007psychological}. 
Despite these consistent findings, their effects are relatively modest, with a recent meta-analysis showing small effect sizes (Pearson \emph{r} of 0.19 for the first-person pronouns, 0.12 for negative emotion words, and -0.21 for positive emotion words)\citep{tolboll2019linguistic}.

Thus far, research has concentrated on identifying the main effects, i.e., the overall significant differences between depression and control groups.
Nevertheless, as can be expected, studies have shown that the linguistic markers of the same person can vary depending on the communication context \citep{mehl2012taking}.

According to Beck's cognitive model of depression \citep{beck1979cognitive,beck2009depression}, schemas of depressive symptoms develop over time and, when active, shape an individual's perceptions, thoughts, and feelings, which influence the linguistic choices of that person when expressing their thoughts and feelings. It is entirely possible and even likely, especially in the case of mild to moderate depression, that the depressive schemas are variably activated in different contexts. Thus, the schema influence on thoughts and linguistic choices are not the same in every context. Therefore, identifying distinct linguistic markers of depression may greatly rely on the context and the activation of depressive schemas at the moment. 
In light of these considerations, a more fine-grained analysis focusing on interaction effects is indicated, considering in which linguistic contexts these differences occur more strongly and which linguistic features co-occur consistently. 

Only a few studies have explored the contextual effect on language markers for depression.
\citet{bernard2016depression} found that higher depression levels correlate with increased use of first-person pronouns. However, they noted that the frequent use of negative emotion words is linked more to higher negative affect than to the depression level itself, suggesting that the prevalence of negative emotion words is not depression per se but rather a negative effect on the state, which is presumably more common in individuals with depression than in healthy controls.

\citet{seabrook2018predicting} and \citet{teodorescu2023language} studied the association between depression and the variability of emotion word rates over time, assuming that people might express different proportions of positive and negative emotion words over time. Both studies found that negative emotion variability was positively associated with depression. That means, two people might express the same overall proportion of negative emotion words, but the higher variability over time (e.g., a higher rate of negative emotion words on one occasion and a lower rate on another occasion) predicted higher depressive symptomatology \citep{seabrook2018predicting} or the diagnostic status \citep{teodorescu2023language}. 

From the computational linguistics point of view, \citet{ireland2018within} studied the linguistic markers of anxiety in 
posts originating from anxiety-related and neutral sub-reddits and found that anxious individuals' word choices differed depending on the sub-reddit.
However, we are unaware of any computational linguistics study that has researched the contextual aspects of linguistic markers of depression.

In this paper, we study the linguistic variation that relates contexts to differential affective tones. Our focus is specifically on understanding the features of the so-called ``depressive language'', i.e., the social science aspect, and not on predicting the diagnostic labels based on textual data, which has been the predominant goal in many previous studies \citep{guntuku2017detecting}.
This approach will help us better understand the varied and context-dependent ways depression influences how people express themselves.

We start with the assumption that the affective quality expressed in texts varies between contexts and thus also necessitates for the authors making different linguistic choices.
We operationalize affective tone as sentiment and contexts as different discussion topics, assuming that some topics activate the depressive schemas more than others. To study the interactions, we use simple linear regression analysis to understand which psycholinguistic features, extracted with the LIWC tool \citep{pennebaker2015development}, correlate with the differential affective tones between depressed and control users over different topics as context.
For our study, we utilize a Reddit-based dataset containing posts from various subreddits of users with and without self-declared depression diagnosis for clinical and control groups, respectively.

Our study centers on the following key research question: 
\noindent
\paragraph{RQ:} 
\emph{\textbf{Which psycholinguistic features affect the sentiment differences observed between depression and control groups across various topics?}}
\paragraph{}

To address this RQ effectively, we begin with a preliminary analysis to lay the groundwork for a deeper inquiry, i.e., \emph{\textbf{Are there differences between depression and control groups regarding the topics discussed and sentiments expressed within our dataset?}}

Based on prior research, we expected that individuals with depression will exhibit more negative sentiment and less positive sentiment \citep{rude2004language,savekar2023structural}. 
Our analysis showed that the posts of people from the depression group showed more negative sentiment. However, contrary to our expectations, we found that the depression group also had more posts with positive sentiment. 
We then followed up with the analyses investigating the RQ. We found that although overall, the depression group used more negative emotion words than the control group, in the contextual analyses, the usage of anger- and sadness-related words were significantly related to the overall positive sentiment of the posts. A small-scale qualitative analysis suggested that posts containing mixed emotions (i.e., references to both positive and negative aspects) might play a role in this correlation. We conclude the paper with some further analyses investigating this direction.

\section{Method}
\subsection{Data}
\begin{table}[h]
\centering
\small
  \begin{tabular}{llll}
       \toprule
       & \bf Depression & \bf Control & \bf Total \\
       \midrule
      \#users & 1316 & 1316 & 2632 \\
      Total \#posts &195.2K &364.4K  & 559.6K \\
      Avg \#posts  & \multirow{2}{*}{148 (78)} & \multirow{2}{*}{277 (146)} & \multirow{2}{*}{213 (133)}\\
      per user (std) & \\
      Avg \#words & \multirow{2}{*}{35 (38)} &\multirow{2}{*}{25 (30)} & \multirow{2}{*}{28 (33)}\\
     per post (std) & \\
       \bottomrule
       
    \end{tabular}
    \caption{Statistics of the balanced depression dataset.
   }
    \label{tab:Dataset}

\end{table}
We used an existing Reddit-based data set comprising posts from users with self-reported mental health diagnoses (SMHD) \cite{SMHD2018}. 
Each diagnosed user is matched with nine control users on average. The data set covers nine mental health diagnoses, including depression.
For this study, we used the depression part only involving 1316 users with the self-reported depression diagnosis. 
We randomly sampled an equivalent number of 1316 control users to create a balanced data set. Additionally, we filtered out all posts containing less than three or more than 200 words. Table~\ref{tab:Dataset} shows the statistics of the study dataset.
More details about the underlying SMHD dataset can be found in Appendix \ref{appendix:Dataset}.

\subsection{Sentiment Analysis}
We evaluated two sentiment models for applicability to our dataset: a RoBERTa-based model, trained on Twitter \cite{barbieri-etal-2020-tweeteval}, and the lexicon-based VADER \cite{hutto2014vader}.
On comparing both models on a set of manually annotated 200 posts randomly drawn from our dataset, we found that although VADER demonstrated slightly higher accuracy (0.69) than RoBERTa (0.66), it more often confused posts with positive and negative sentiments. Therefore, we chose the RoBERTa-based model for subsequent analyses.
For more details of the comparative analysis, refer to the Appendix \ref{appendix:sentiment}.

\subsection{Topic Modeling}
BERTopic \cite{grootendorst2022bertopic} leverages the power of transformer-based language models to capture the contextual information and meaning of words in each document. 
We used the default topic model setting for our purpose, as according to the BERTopic documentation,\footnote{https://maartengr.github.io/BERTopic/index.html} the default model works quite well for most use cases. More than 5000 topics were initially derived from the model, with about 50\% of the data classified as outliers (documents not fitting any topic, labeled as -1). Utilizing the ``reduced outlier'' function, which leverages the c-TF-IDF strategy and cosine similarity, the proportion of outliers was reduced to 0.25\%. Additionally, the ``automatic topic reduction'' function was applied to reduce the number of topics, resulting in 4187 topics.

\subsection{LIWC Analysis}
LIWC (Linguistic Inquiry and Word Count) is a lexicon-based tool that analyzes texts by mapping words to psycho-linguistic attributes, resulting in the proportion of words in various categories. These categories include Summary Variables, Linguistic Dimensions, Psychological Processes, and more, detailed in \cite{pennebaker2015development} and Appendix \ref{appendix:LIWC}. Among the 110 LIWC attributes, we selected 63 attributes relevant to our research objectives that were not highly correlated to each other to avoid multicollinearity in the subsequent linear regression analysis.
We assessed the correlations using Pearson's correlation coefficient with a threshold of 0.5. Appendix~\ref{appendix:LIWC} shows the list of selected attributes.

\subsubsection{User-based LIWC Analysis} Because our data is a balanced and length-restricted subset of the SMHD depression dataset, we first analyze the difference in LIWC attributes between the depression and control groups and compare our results to those reported by \citet{SMHD2018}. Similarly to the cited study, the group means are aggregated over users, i.e., for computing the feature values for a single user, all their posts were first concatenated. Similarly, we performed Welch's t-test \cite{welch1947generalization} with adjusted p-value using Bonferroni correction. For effect size assessment, we calculated Cohen's d statistics \cite{cohen1987statistical}.

\subsubsection{Topic-Specific LIWC Analysis} 
\label{sec:ols_analysis}
In the main analysis of the paper, we wanted to understand which LIWC attributes affect the sentiments expressed in relation to various topics. Moreover, we wanted to capture the sentiment differences between depression and control groups. For that, we employed linear regression analysis with the features derived from LIWC attributes as independent variables and an overall sentiment polarity difference between groups as the dependent variable.
\begin{table}[b]
\centering
    \small
   \addtolength{\tabcolsep}{-0.1em}
   \begin{tabular}{lccccc}
        \toprule
         Topics&  \multicolumn{2}{c}{Depression}& \multicolumn{2}{c}{Control}& Sent diff\\ 
         \cmidrule(lr){2-3}
         \cmidrule(lr){4-5}
         \cmidrule(lr){6-6}
          & pos&  neg&   pos&  neg& y\\ 
         \midrule
         0 Animals&28.0&26.0&19.1&23.9&  6.8\\ 
         1 Relationships& 14.4&46.4&13.4&31.3& -14.1\\ 
         2 US elections& 20.2&18.5&16.8&73.4&-5.0 \\ 
         \bottomrule
    \end{tabular}
    \caption{Dependent variable calculation.}
    \label{tab:y_cal}
\end{table}
\paragraph{The dependent variable,} i.e., the sentiment difference between depression and control groups per topic, is calculated as the net sentiment score difference denoted as $y = (pos - neg)_{\text{depression}} - (pos - neg)_{\text{control}}$ (see Table \ref{tab:y_cal} for some examples), where pos and neg columns show the percentage of positive and negative sentiments per topic and group, respectively.
For instance, in Topic 0 (Animals), for the depression group, 28.0\% of posts are labeled as positive, whereas 26.0\% have a negative sentiment. The net difference is 2\%, showing that overall, the depression group has slightly more positive sentiment towards that topic. In contrast, for the control group, the net difference is $19.1\%-23.9\%=-4.8\%$, showing that overall, the control group has more negative sentiment towards that topic. Subtracting these differences $y=2-(-4.8)=6.8$ yields an outcome value capturing the overall difference between depression and control groups towards that topic. Positive difference refers to more positive sentiment in the depression group posts, while negative value means more positive sentiment in the control group posts. Values close to zero indicate the similarity of positive and negative sentiment proportions in both depression and control groups.  

\paragraph{Independent features} were computed in two steps. First, we calculated mean aggregated LIWC attribute scores for both groups topic-wise. For instance, consider Topic 0 (Animals) and the Analytic feature (see Table~\ref{tab:table_features}). For the depression group, the average score of that attribute for Topic 0 is $43.0$, while for the control group it is $49.7$. Subsequently, we calculated the difference between these scores, i.e., $f_{\text{Analytic}} = 43.0 - 49.7 = -6.7$. Positive feature values refer to higher proportion of the attribute value in the depression group, while negative feature values mean that control group had more of that attribute.
Independent features were computed this way for all topics and 63 selected attributes, resulting in a size $4187 \times 63$ (topics $\times$ attributes) matrix. This dataset was used to fit the linear regression model using the ordinary least squares method.
\begin{table}[h]
    \small
    \addtolength{\tabcolsep}{-0.1em}
    \begin{tabular}{lcccccc}
        \toprule
         Topics&  \multicolumn{3}{c}{Analytic}&  \multicolumn{3}{c}{Clout}\\ 
         \cmidrule(lr){2-4}
         \cmidrule(lr){5-7}
          & Dep&  Ctr& $f_A$ &  Dep&  Ctr& $f_C$\\ 
         \midrule
         0 Animals&  43.0&  49.7&  -6.7&  47.3&  49.6& -2.3\\ 
         1 Relationships&  25.5&  26.5&  -1.0&  59.3&  62.3& -3.0\\ 
         2 US elections&  48.8&  57.2&  -8.4&  32.4&  34.2& -1.8\\ 
         \bottomrule
    \end{tabular}
    \caption{Independent feature calculation for the regression analysis (Dep = Depression group, Ctr = Control group,
    $f_A =f_{\text{Analytic}}$ ,$f_C = f_{\text{Clout}}$).}
    \label{tab:table_features}
\end{table}

\section{Results}
\subsection{User-based LIWC analysis}
The analysis, shown in Table~\ref{tab:LIWC_result_1}, revealed more significant differences and larger effect sizes between the groups than those reported by \citet{SMHD2018}. This discrepancy may arise firstly because we only tested 63 pre-selected, uncorrelated attributes, which makes the adjusted p-value threshold higher than it would be with the full attribute set used by \citet{SMHD2018}. Secondly, although \citet{SMHD2018} does not report means and standard deviations, we expect the standard deviations to be smaller in our subset due to restrictions on the post length, which affects both the attribute significance and the magnitude of the effect sizes.
 \begin{table*}[ht!]
        \small
        \centering
        \begin{tabular}{lSSSScSr}
            \toprule
            \multirow{2}{*}{\bf LIWC Attributes} & \multicolumn{2}{c}{\bf Depression} & \multicolumn{2}{c}{\bf Control} & {\bf p-value} & {\bf Cohen's d} & {\bf Cohen's d}\\
            & {mean} & {std} & {mean} & {std} & & {this study} & {\citet{SMHD2018}} \\
            \midrule
            Word Count  & 36.1 & 14.5 & 24.6 & 10.9 & *** & 0.90 & N/S  \\
            Analytic    & 42.4 & 9.9  & 48.8 & 10.0 & *** &-0.63 & N/S  \\
            Clout       & 37.5 & 10.8 & 40.6 & 9.7  & *** &-0.30 &-0.06 \\
            Authentic   & 56.0 & 9.7  & 49.6 & 9.7  & *** & 0.65 & 0.2  \\
            \midrule
            1st person singular & 6.1 & 1.8 & 4.7 & 1.8 & *** & 0.77 & 0.23 \\
            3rd person singular & 1.2 & 0.9 & 1.0 & 0.7 & *** & 0.29 & 0.09 \\
            Impersonal pronouns & 5.8 & 0.9 & 5.5 & 1.2 & *** & 0.22 & 0.06 \\
            \midrule
            Insight             & 2.8 & 0.7 & 2.4 & 0.7 & *** & 0.49 & 0.09 \\
            Causation           & 1.7 & 0.4 & 1.6 & 0.5 & **  & 0.15 & N/S  \\
            Tentative           & 3.1 & 0.7 & 2.9 & 0.8 & *** & 0.28 & 0.07 \\
            Certitude           & 0.9 & 0.3 & 0.8 & 0.3 & *** & 0.34 & N/S  \\     
            Differentiation     & 3.7 & 0.7 & 3.6 & 0.9 & *** & 0.19 & 0.08 \\
            Negative emotion    & 0.7 & 0.3 & 0.5 & 0.3 & *** & 0.65 & N/S  \\
            Anxiety             & 0.1 & 0.1 & 0.1 & 0.1 & *** & 0.60 & 0.07 \\
            Anger               & 0.2 & 0.1 & 0.1 & 0.1 & **  & 0.16 & N/S  \\
            Sadness             & 0.1 & 0.1 & 0.1 & 0.1 & *** & 0.44 & N/S  \\
            Swear words         & 0.4 & 0.4 & 0.6 & 0.7 & *** &-0.24 & N/S  \\
            Social behavior     & 3.7 & 0.9 & 3.5 & 0.9 & *** & 0.26 & N/S  \\
            Interpersonal       & 0.3 & 0.2 & 0.3 & 0.2 & *   &-0.15 & N/S  \\
            Communication       & 1.7 & 0.6 & 1.5 & 0.6 & *** & 0.37 & N/S  \\
            Family              & 0.4 & 0.4 & 0.2 & 0.3 & *** & 0.40 & 0.06 \\
            Friends             & 0.2 & 0.2 & 0.2 & 0.2 & *** & 0.25 & N/S  \\
            Female references   & 0.8 & 0.7 & 0.5 & 0.6 & *** & 0.45 & 0.13 \\
            \midrule
            Culture         & 0.7 & 0.5 & 0.9 & 0.9 & *** &-0.35 & N/S  \\
            Lifestyle       & 3.0 & 1.0 & 3.3 & 1.2 & *** &-0.27 & N/S  \\
            Illness         & 0.2 & 0.2 & 0.1 & 0.1 & *** & 0.54 & N/S  \\
            Wellness        & 0.1 & 0.1 & 0.1 & 0.1 & *** & 0.26 & N/S  \\ 
            Mental health   & 0.1 & 0.1 & 0.02& 0.07& *** & 0.50 & N/S  \\
            Substances      & 0.1 & 0.2 & 0.1 & 0.1 & *** & 0.21 & N/S  \\
            Want            & 0.4 & 0.2 & 0.3 & 0.2 & *** & 0.29 & N/S  \\
            Fatigue         & 0.04& 0.05& 0.03& 0.07& *   & 0.15 & N/S  \\
            Reward          & 0.1 & 0.1 & 0.2 & 0.2 & *** &-0.28 & N/S  \\
            Risk            & 0.3 & 0.1 & 0.2 & 0.1 & *** & 0.25 & N/S  \\
            Perception      & 8.3 & 1.2 & 8.5 & 1.6 & **  &-0.16 & N/S  \\
            Feeling         & 0.5 & 0.3 & 0.4 & 0.3 & *** & 0.42 & N/S  \\
            Past focus      & 3.6 & 1.0 & 3.2 & 1.0 & *** & 0.41 & 0.08 \\
            Present focus   & 3.9 & 0.8 & 4.1 & 1.0 & *** &-0.27 & 0.09 \\
            Conversational  & 1.0 & 0.6 & 1.3 & 0.9 & *** &-0.38 & N/S  \\
            \bottomrule  
        \end{tabular}
        \caption{Difference between psycholinguistic LIWC attributes between depression and control groups. Only statistically significant attributes are shown. Adjusted p-value thresholds for multiple comparisons is 0.00079 and denoted as * $<$ 0.00079, ** $<$ 0.00015, *** $<$ 0.000015, equivalent to * $<$ 0.05, ** $<$ 0.01, *** $<$ 0.001. The table includes effect sizes (Cohen's d) for both this study and \citet{SMHD2018}, with `N/S' indicating no statistical difference found in the latter study.
        }
        \label{tab:LIWC_result_1}
    \end{table*}

Most attributes that showed statistical significance in both studies (13 out of 14) had consistent effect sizes, except for the present focus. While \citet{SMHD2018} found a small positive effect size, indicating a higher proportion of present-focused words in the depression group, our subsample showed a negative effect size, aligning with studies suggesting that individuals with depression are generally less present-focused \citep{nolen2008rethinking, rodriguez2010reading}. Thus, our random subsample appears to be representative of the whole depression part of the SMHD data.

\subsection{Sentiment Analysis}
\label{sec:Sentiment}
\begin{figure}[t]
\centering
\includegraphics[width=0.45\textwidth]{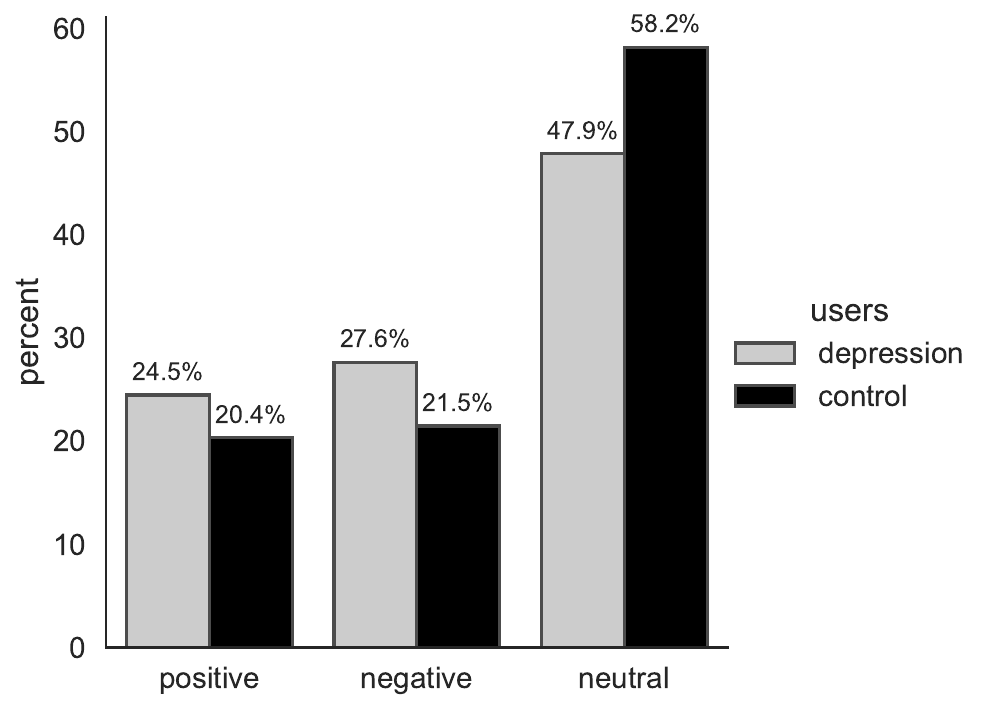}
\caption{Sentiment distribution of both groups.}
\label{Fig:Sentiment}
\end{figure}

As shown in Figure~\ref{Fig:Sentiment},
the depression group exhibits more negative sentiments (6\%) and, unexpectedly, more positive sentiments (4\%) than the control group. This variation in sentiment distribution is statistically supported by the Chi-square test ($\chi^2(2)= 5503.79, p < 0.001$).
This suggests that the users with depression not only express more negative emotionality but display an overall higher emotionality in their posts.

\subsection{Topic Modeling}
Topic modeling resulted in 4187 topics discussed by both groups, of which 13 low-frequency topics were unique to one group only. These topics and their posts were removed from the data, resulting in 4174 topics. 
We analyzed the frequency and sentiment distribution of these topics among the user groups. Despite the general common topic discussions, noticeable differences in topic prevalence and sentiment patterns emerged between the depression and control groups. 
While the detailed examination of topic modeling outputs is not the central focus of our research, as its purpose is to segment the data into contexts for further analyses, the differences in topic distributions and sentiment distributions in topics between groups might be interesting in their own right. We provide some further analyses in~Appendix \ref{appendix:Topicmodeling}.

\subsection{Topic-Specific LIWC Analysis}
\label{sec:topic_wise_LIWC}
\begin{figure}[t]
\centering
    \includegraphics[width=0.48\textwidth]{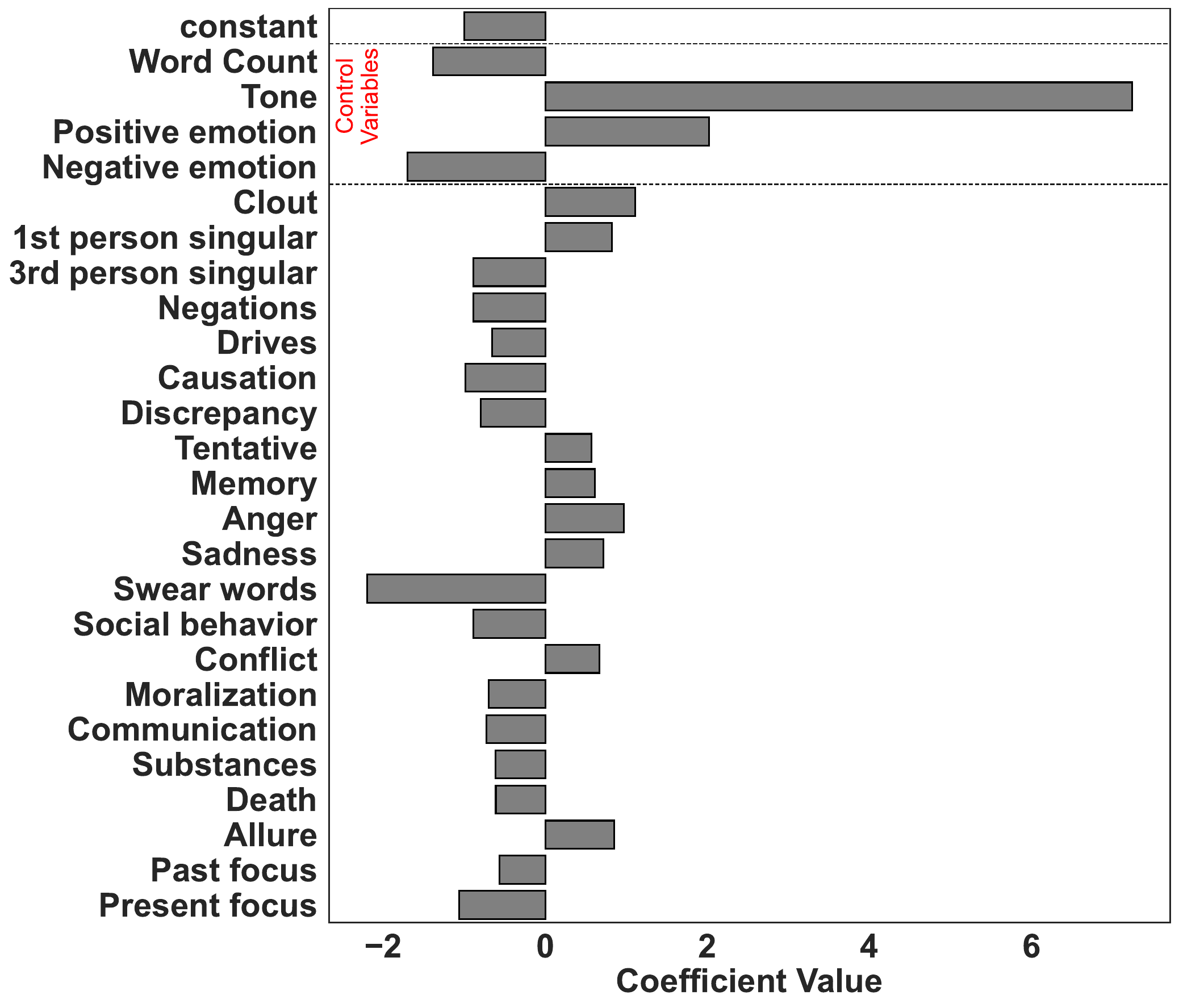}
    \caption{The linear regression model coefficients of the statistically significant features and their impact on the sentiment difference.}
    \label{Fig:ols_re}
\end{figure}

Using a linear regression model with 63 LIWC attributes as predictors, we assessed the sentiment differences outlined in Section~\ref{sec:ols_analysis}. The model accounted for 26.6\% of the variability in sentiment difference, $R^2 = 0.266, F(63,4110) = 23.63, p < 0.001$, with 25 attributes being statistically significant ($p = 0.05$). 
Figure~\ref{Fig:ols_re} illustrates the impact of the significant attributes on the sentiment difference outcome variable. Recall that positive attribute values refer to the depression group expressing more of that attribute in a topic. Thus, positive model weights indicate these features are associated with higher positive sentiment in the depression group posts across topics. Complete results are in Appendix \ref{appendix:OLS}.

The bias term was significant, and its negative value shows that overall, the sentiment of the depression group tends to be more negative than that of the control group.  
Control variables such as word count, tone, and positive and negative emotions showed expected correlations with sentiment values. The average post length in our data was longer for the depression group compared to the control group (see Table~\ref{tab:Dataset}). The negative coefficient shows that higher positive sentiment for the depression group is associated with shorter posts than the control group.
The positive and negative emotion features are expected to be closely correlated to the positive and negative sentiment values. The coefficients of these features are in the expected direction. Finally, the tone feature, which includes both positive and negative tone, has the largest absolute coefficient and a positive weight, having the largest impact on the outcome variable. 

A higher proportion of first-person singular pronouns is related to a higher positive sentiment, and a higher proportion of third-person singular pronouns is related to a higher negative sentiment. Time orientation features, such as past focus and present focus, have negative coefficients, meaning that a higher proportion of words in those categories are related to a higher negative sentiment.
There are some features that, with their negative weight, are perhaps quite expectedly related to a higher negative sentiment: swear words, moralization, substances, and death. In contrast, more conflict-related words are associated with higher positive sentiment. 

Finally, we highlight the two remaining emotion-related features: anger and sadness. These features have positive weights, correlating with more positive sentiment. In contrast to other findings mentioned above, these results are surprising, as anger and sadness as instances of negative emotions are expected to be more correlated with overall negative sentiment. In the next section, we will attempt to understand these findings. 

\begin{table*}[ht!]
    \centering
    \small
    
    \begin{tabular}{p{0.2cm}p{1.0cm}p{10.0cm}p{1.1cm}p{1.4cm}} 
        \toprule
         \bf No & \bf Topic & \bf Post text & \bf Sentiment & \bf Attribute \\
        \midrule
         1 & Art & I truly admire and appreciate the art; it's impressive. Yet, it's also causing me a great deal of frustration. It's amazing, though.	& Positive & Anger \\
         \midrule
         2 & Empathy & In times of sadness, we seek understanding and compassion. It's music that has the power to uplift our spirits. & Positive  & Sadness\\ 
         \midrule
         3 & Albums & My all-time favorite musical work is the second symphony. It deeply saddens me. & Positive & Sadness\\
         \midrule
         4 & Animals & Whenever I'm feeling low, a walk with my dog always helps. He invariably does something silly or amusing during our walk, which never fails to lift my spirits. & Positive & 1st person pronoun \\
         \midrule
         5 & Gym & Previously, I relied on gym buddies, but their absence meant I stopped too. Now, I've taken control—working out alone, focusing on my diet, and tracking my progress. Sometimes friends join, but mostly, it's just me. This self-reliance has led to sustained success for the first time. My motivation and achievements are my own, though I welcome occasional companionship and encouragement. This self-empowered approach has transformed into my lifestyle, leaving no room for excuses, for myself or others. & Positive & 1st person pronoun\\
         
         \midrule
         6 & Family & My father left when I was a child, leaving me confused about his reasons. Over time, living with my mother helped me understand his choice, though being with her has been challenging. I wish he hadn't left on my birthday. Despite this, we've reconnected and improved our relationship. & Negative & Past focus \\
         \midrule
         7 & Emotions & Do you ever worry that just as life gets better, something bad will happen? This fear of sudden, negative changes when things are going well makes me hesitant to fully invest myself. How do you deal with this anxiety? & Negative & Present focus \\
         \bottomrule       
    \end{tabular}
    \caption{Posts of depressed users with sentiment label and relative language attributes and associated topics. All posts have been rephrased to maintain the privacy of users.}
    \label{tab:posts}
\end{table*}

\section{Anger and Sadness}
\label{sec: mixed emo}

In the previous section, we found that the use of anger- and sadness-related words, typically seen as negative emotion words, correlates with positive sentiment. 
In order to get some idea of the observed phenomenon, we reviewed some of the posts with positive sentiments that contained anger- and sadness-related words. Examples 1--3 shown in Table~\ref{tab:posts} suggest that these posts express what might be called \emph{mixed emotions}, i.e., containing a mixed usage of positive and negative emotion words.

Building on this observation, we designed two analyses to study 1) if posts with mixed emotions could be responsible for overall higher positive sentiments observed for the depression group, and 2) if posts with mixed emotions might significantly contribute to the observed positive relationship between the anger- and sadness-related feature and positive sentiment. 

\paragraph{Mixed Emotions:} First, we need to operationalize what it means for a post to display mixed emotions. We define posts with mixed emotions as those that contain both positive and negative emotion words, i.e., the LIWC attribute of both positive and negative emotions is greater than zero. 
We aimed to examine the role of mixed emotion posts in our findings by comparing subsets of data both with and without mixed emotions.
Given the possible combination of emotion words, we further categorized the data into four segments rather than with or without mixed emotions, which are: Mixed Emotions (both positive and negative emotion words, 3.6\% of the total posts), Positive Emotions (only positive emotion words, 15.9\%), Negative Emotions (only negative words, 8.7\%), and Neutral Emotions (neither positive nor negative words, 71.8\%). 

\paragraph{Sentiment Distribution:}
\begin{figure}[t]
\centering
\includegraphics[width=0.48\textwidth]{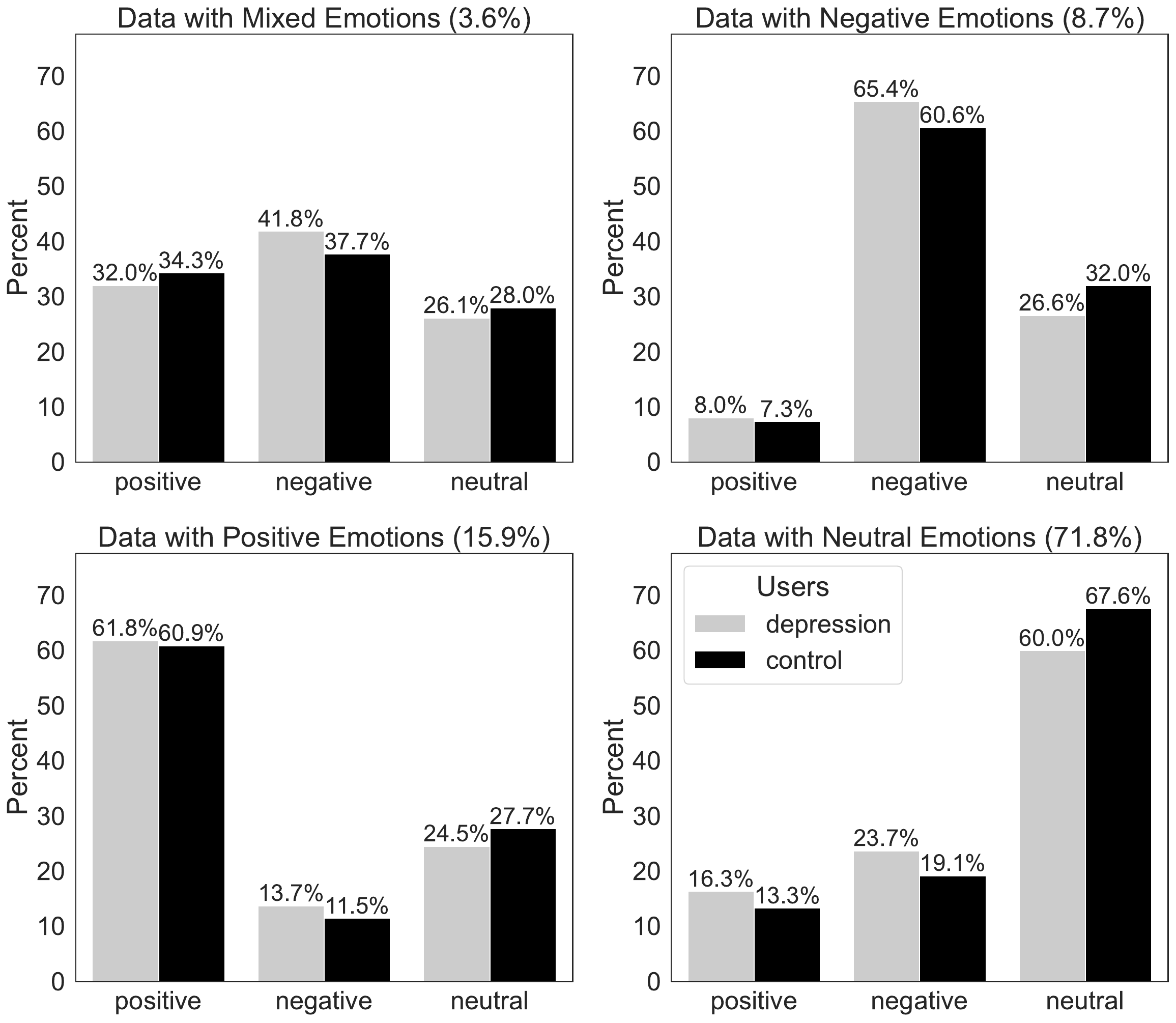}
\caption{Sentiment distribution across data splits.}
\label{Fig:senti_data}
\end{figure}

First, we looked at the differences in sentiment proportions between depression and control groups in each of those data splits (shown in Figure~\ref{Fig:senti_data}). 

As expected, negative sentiments dominate in Negative Emotions split, positive sentiments in Positive Emotions split, and neutral sentiment in Neutral Emotions split. 
Mixed Emotions data shows an almost uniform distribution over sentiments. 
When looking at group differences in the positive sentiment, the proportion is similar in both groups in both the Mixed Emotions, Positive Emotions and Negative Emotions split, while in the Neutral Emotions group the depression group has more positive sentiment. Because the Neutral split is the biggest (72\%), we conclude that this split, instead of the Mixed Emotions as we expected, drives the overall sentiment pattern observed in Figure~\ref{Fig:Sentiment}. 
This analysis highlights an important limitation of lexicon-based systems, which struggle to grasp the overall context and sentiment in the absence of polarized emotional words. While the RoBERTa-based sentiment model is not perfect, it can capture emotional tone that is concealed from the lexicon-based LIWC system. 

\begin{figure}[t]
\centering
\includegraphics[width=0.48\textwidth]{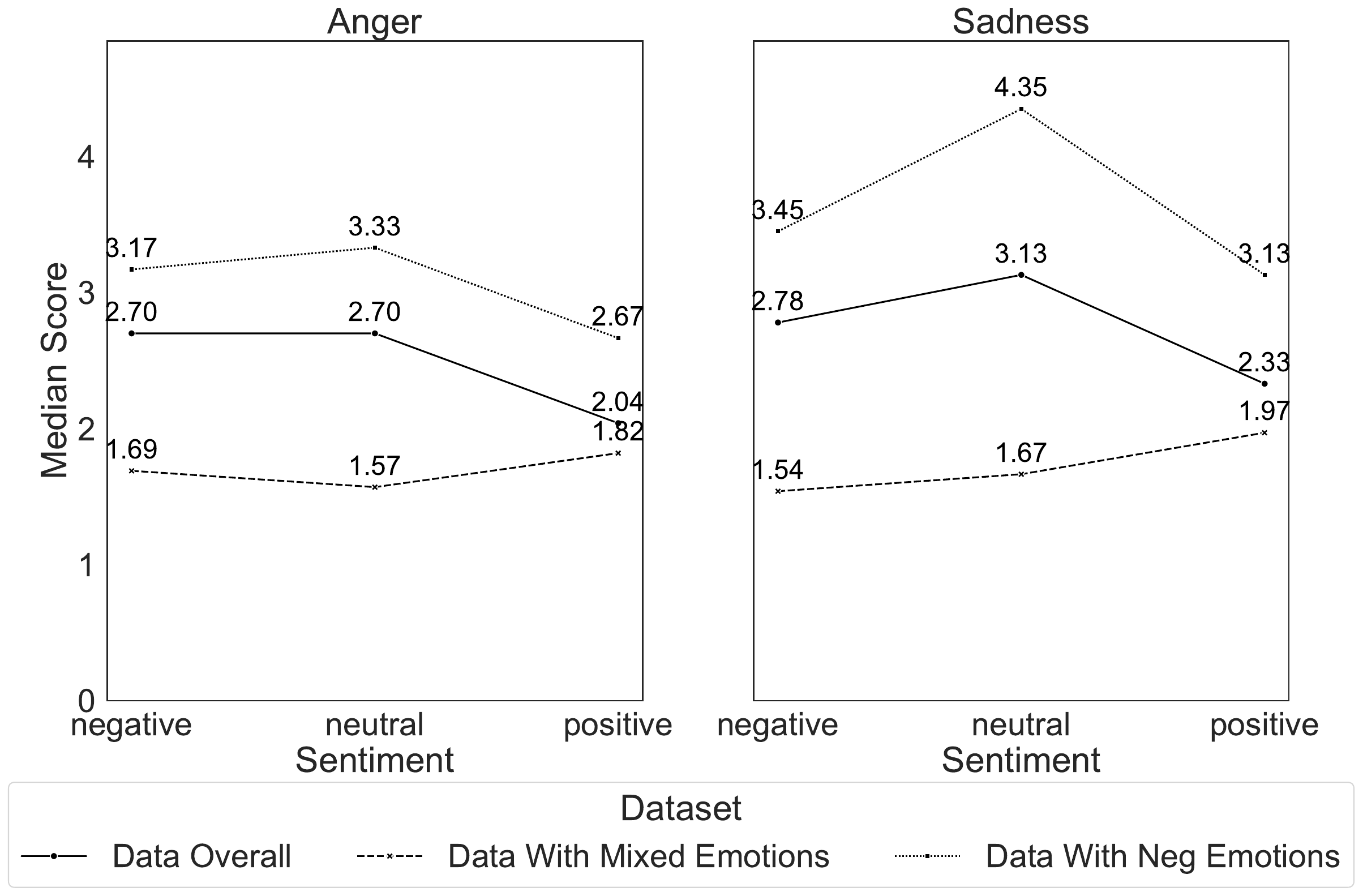}
\caption{Median anger and sadness scores across data splits.}
\label{Fig:anger_sadness}
\end{figure}

\paragraph{Anger and Sadness in Mixed Emotions:}
Next, we explored if mixed emotion posts might be related to positive correlations between anger and sadness features and the positive sentiment as found in Section~\ref{sec:topic_wise_LIWC}.
Figure~\ref{Fig:anger_sadness} plots the median anger and sadness scores for the overall data, Mixed Emotions split and the Negative Emotions split.\footnote{Positive and Neutral Emotions splits are omitted as by definition they do not exhibit any negative emotion words.}
Both anger and sadness scores were highest in the Negative Emotions split and lowest in the Mixed Emotions split across all sentiments.
However, the overall pattern of median anger and sadness scores in Mixed Emotions split differs from other splits. In contrast to other slits, both the median anger and sadness scores are \emph{highest} in positive sentiment posts.

\section{Discussion}

This study aimed to investigate the relationship between (psycho)linguistic features and affective tone across contexts operationalized as topics in depressed and non-depressed reddit users.
Overall sentiment analysis revealed that the depression group expressed more negative sentiment, aligning with prior studies  \citep{liu2022relationship}, yet also more positive sentiment, which was unexpected. 
In further analyses that split the data into four subsections regarding the presence or absence of positive or negative emotion words, we found that the so-called neutral posts containing neither positive nor negative emotion words according to LIWC were driving this pattern.
This finding highlighting the constraints of the lexicon-based systems, which fail to capture the full emotional context in absence of explicit emotional words. 

A higher proportion of first-person singular pronouns was related to a higher positive sentiment, contrasting with previous research linking it to negative emotionality \citep{tackman2019depression,bernard2016depression}. 
However, another study \citet{brockmeyer2015me} found that for people with depression, the higher usage of first-person singular pronouns was related to texts elicited in the positive memory recall task but not in the negative memory recall task. Thus, although the majority of studies \citep{lyons2018mental, rude2004language, stirman2001word, demunmun2013paper,savekar2023structural,trifu2017linguistic,chung2007psychological} (including this one) have found the overall higher usage of first-person pronouns by individuals with depression, the interaction studies considering either affective, topical, or other contexts might show a different and more interesting picture, the analysis of which could be a topic of future studies.
Examples 4 and 5 in Table~\ref{tab:posts} provide illustration; despite the self-referential nature of the language, the sentiments expressed in these posts are notably positive.

In the main effect LIWC analyses, the depression group displayed more past-focused language and less present-focused language, similar to previous works \citep{trifu2017linguistic,smirnova20131419, imbault2018emotional}. At the same time, in linear regression analyses, both past-focused and present-focused features were negatively associated with the positive sentiment difference. 
As an illustration, in the Example 6 in Table~\ref{tab:posts}, the user expresses negative affect in relation to past experiences. In contrast, in Example 7, the user conveys a current state of worry, leading to negative sentiments about present circumstances, demonstrating that in certain contexts, present focus might be an indication for depressive language.

Finally, we observed a significantly higher use of negative emotional words in the depression group, including anxiety-, anger- and sadness-related words, which is in line with previous studies \citep{trifu2017linguistic,savekar2023structural,rude2004language,chung2007psychological}.
In the linear regression analysis, although the association between the negative emotion words and the sentiment difference were, similarly to the main effect, negative, the correlations between the sentiment difference and the anger and sadness features were positive, i.e., the higher rate of anger and sadness words were related to more positive sentiment. 

Qualitative analysis of positive sentiment posts with non-zero anger or sadness scores revealed a pattern of mixed emotions, i.e., texts containing features of both positivity and negativity, such as starting by describing something negative, but ending in a positive note.
When exploring the potential role of mixed emotions in this relation, we found that in contrast to other types of posts, the posts with mixed emotions have the highest anger and sadness scores in posts with positive sentiment. While this result does not provide definitive evidence for the role of mixed emotions in the observed positive correlations between anger and sadness features and positive sentiment, it shows that posts with mixed emotions behave differently from other posts containing negative emotion words and thus can play a different role in the depressive language.

\section{Conclusion}
In conclusion, our research highlights the important role of discussion context in shaping emotional expressions among individuals with (but also without) depression. Contrary to prior studies, we observed not only more negative sentiments but also more positive sentiments within the depression group---a pattern that was not captured by the LIWC, illustrating the limitations accurately interpreting emotions in the absence of explicit emotional words.
Analyses also revealed notable interactions between linguistic markers—such as anger and sadness—and positive sentiments, suggesting a  potentially important role of posts with mixed emotions. 
In summary, our findings support the notion that the research in linguistic markers of depression requires going beyond studying main effects and necessitates a contextual and multifaceted approach.

\section*{Limitations}
\label{sec:limits}
There are several limitations to consider in our study. Firstly, the validity of the sentiment analysis model cannot be ensured because although the model is trained on social media data (Twitter), our data comprises Reddit posts, which, even with the length restriction imposed on our subsample, are, on average, considerably longer than tweets. We assessed the model's performance by manually annotating a random subset of 200 posts and found that the disagreements stemmed mostly from the model's tendency to categorize positive and negative posts as neutral. Thus, it is likely that the amount of posts with positive and negative sentiments is somewhat underestimated.

Two key limitations concern the SMHD dataset utilized.
Firstly, the dataset spans from 2006 to 2017. It is important to acknowledge that the presence and severity of depression may vary over time for individuals \citep{harrigian2022then}. There is a possibility that some users labeled as depressed did not have depression during the entire timeline, and such temporal uncertainty may impact the interpretation of results. Second, the control group is auxiliary \citep{ernala2019methodological}, i.e., although the control group was selected from non-mental health-related sub-Reddits, there is no way to be sure if controls are actually controls or if there are users in the control group who might be on the spectrum with any mental health disorders. 

The LIWC tool, despite its widespread use, faces limitations due to its lexicon-based approach particularly with contextual nuances. An example includes incorrectly assigning a high anger emotion score to a statement like "Rita Madder is freaking great" because of the word "Madder," despite no anger being expressed. Our findings further validate this limitation, as neutral emotion settings revealed a significant difference in positive sentiment distributions among groups, highlighting the inadequacy of lexicon-based systems in capturing the true sentiment context.

Additionally, our regression analysis explains only 26\% of the variance of the sentiment difference, indicating that unaccounted factors might influence the observed patterns in language and sentiments. This limitation could partly stem from LIWC's challenges in capturing context.
Moreover, our findings are derived from a dataset specifically concerning depression, limiting the generalizability of our conclusions to broader populations and contexts. Further research with diverse datasets is necessary to apply these results more universally.

\section*{Ethical Considerations}
In our study, we analyzed the language of the social media posts of both depressed and non-depressed users. We used the existing SMHD dataset \cite{SMHD2018} that we obtained from its creators by signing a user agreement; we have adhered to the terms and conditions outlined in this agreement when conducting this study. 
In our work, we search for general patterns and do not make predictions or draw conclusions about any particular user in the dataset. Also, we believe that our findings are interesting for the social science sphere, however, we believe that they will not be directly useful for drawing conclusions about users posting in social media.

\section*{Data and Code Availability Statement}
In the interest of fostering transparency and reproducibility, the source code supporting the findings of this study is publicly available. The code repository, which includes the scripts and any additional documentation necessary for replicating the analyses and results presented in this paper, can be accessed at the following GitHub link.\footnote{https://github.com/nehasharma666/Depression}. For access to the data itself, please contact the authors of \citet{SMHD2018}.

\section*{Acknowledgement}
This research was supported by the Estonian Research Council Grant PSG721.

\bibliography{acl_latex}

\appendix

\section*{Appendices}
\section{Dataset}\label{appendix:Dataset}
SMHD is a collection of self-reported mental health diagnoses from Reddit, designed for academic and research purposes  \cite{SMHD2018}. 
SMHD contains posts of Reddit users with nine mental health disorders along with matched control users from a period spanning from January 2006 to December 2017 including depression, ADHD, anxiety, bipolar, PTSD, autism, OCD, schizophrenia, and eating disorder.  

The SMHD dataset includes posts from 20,406 clinical users who have claimed to have been diagnosed with a mental health condition and 335,952 control users who are unlikely to have one of the mental health conditions studied. The clinical users were identified based on the textual patterns of self-reported diagnosis (e.g., I was diagnosed with depression) and keywords related to diagnoses (language related to mental health such as the name of a condition, and general terms like diagnosis, mental illness, or suffering from, etc.). Control users were selected based on the criteria that they had not posted in any mental health related subreddits. Control users were selected from a group of potential candidates based on their similarity to clinical users, determined by their subreddit activity and number of posts. The criteria for selecting control users were rigorous: candidates were excluded if they did not meet the required subreddit overlap or minimum post count criteria or used any mental health-related terms in their posts.

After collecting all related user posts, the authors of \citet{SMHD2018} removed all mental health-related posts for clinical users to make the data of both user groups similar. In the dataset, each user is represented by a unique identification number, and their data includes the text of the posts made by that user. Table \ref{tab:SMHD_data} shows the number of posts and tokens per diagnosis.

\begin{table}[ht!]
\centering
\begin{tabular}{lrr}
\toprule
\textbf{Diagnosis}& \textbf{Posts} & \textbf{Tokens} \\
\midrule
depression  & 1,272K& 57.4M  \\
adhd        & 872K  & 40.5M \\
anxiety     & 795K  & 36.9M  \\
bipolar     & 575K  & 26.2M   \\
ptsd        & 258K  & 13.7M  \\
autism      & 248K  & 11.6M   \\
ocd         & 203K  &  9.4M\\
schizophrenia&123K  & 6.1M \\
eating      & 53K   & 2.5M \\
\midrule 
control     &115,669K& 3,031.6M  \\
\bottomrule 
\end{tabular}
\caption{SMHD dataset statistics.}
\label{tab:SMHD_data}
\end{table}

\section{Sentiment Analysis}\label{appendix:sentiment}
We evaluated two sentiment analysis models for applicability to our dataset: a transformer-based model, RoBERTa, trained on Twitter \citep{barbieri-etal-2020-tweeteval}, and the VADER sentiment analysis model \citep{hutto2014vader}. After manually annotating 200 posts for ground truth by the first author of this paper, we compared both models' predictions against these annotations.

\begin{table*}[ht!]
    \centering
    \begin{tabular}{lcccccc}
    \toprule
     & \multicolumn{2}{c}{Precision} & \multicolumn{2}{c}{Recall} & \multicolumn{2}{c}{F1}\\   
    \cmidrule(lr){2-3}
    \cmidrule(lr){4-5}
    \cmidrule(lr){6-7}
    & Vader & RoBERTa & VADER & RoBERTa & VADER & RoBERTa \\
    \midrule
    Positive & 0.64 & 0.71 & 0.68 & 0.75 & 0.66 & 0.73\\
    Negative & 0.81 & 0.53 & 0.50 & 0.79 & 0.62 & 0.64\\
    Neutral  & 0.66 & 0.88 & 0.85 & 0.48 & 0.74 & 0.62\\
    \midrule 
    Accuracy    &      &      &      &      & 0.69 & 0.66\\
    Macro       & 0.70 & 0.71 & 0.68 & 0.68 & 0.67 & 0.67\\
    Weighted    & 0.71 & 0.72 & 0.69 & 0.66 & 0.68 & 0.66\\    
    \bottomrule
    \end{tabular}
    \caption{Classification Report for the VADER and RoBERTa sentiment models.}
    \label{tab:classification}
\end{table*}

\begin{figure*}[ht]
    \includegraphics[width=0.48\textwidth]{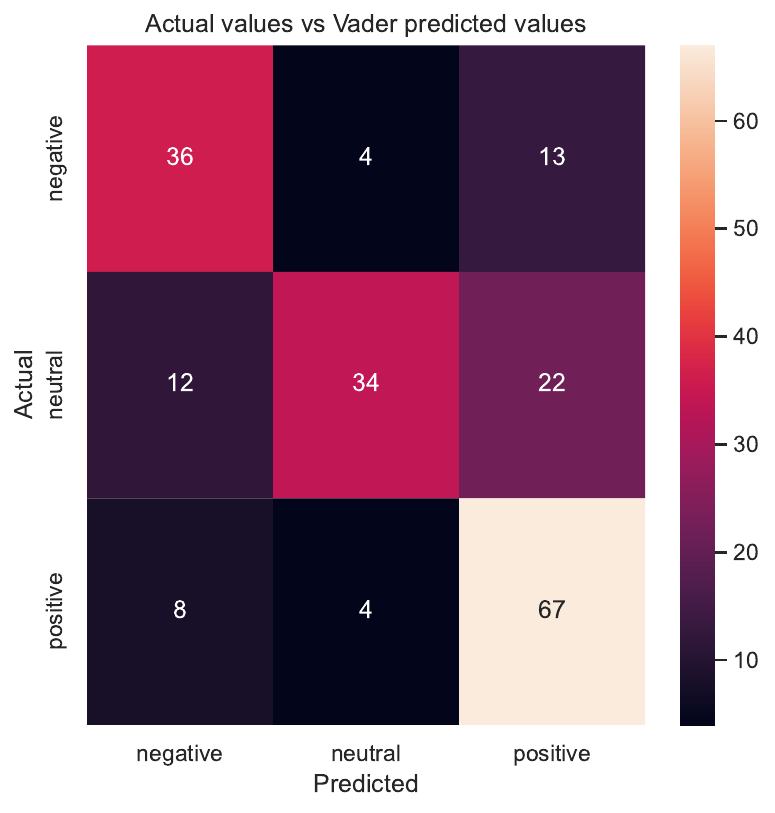}
    \includegraphics[width=0.48\textwidth]{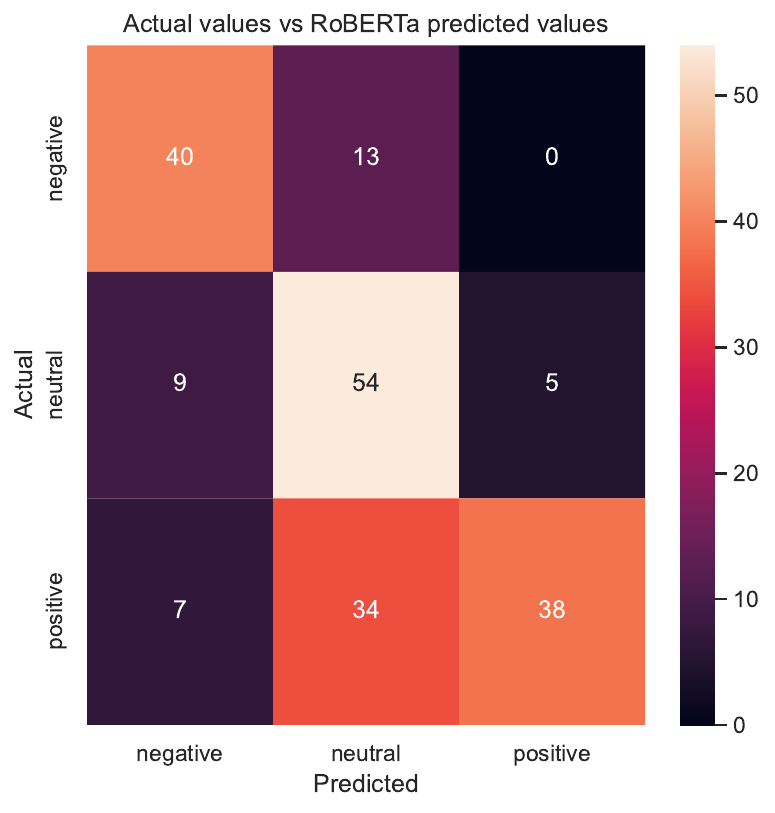}
    \caption{Confusion matrices of the VADER and RoBERTa predicted values compared to the manually annotated subset of 200 posts.}
    \label{fig:sentiment_confusion}
\end{figure*}

The VADER model exhibited a slightly higher overall accuracy of 69\% compared to the 66\% achieved by the transformer-based model (see Table \ref{tab:classification}). At the same time, compared to the RoBERTa model, VADER shows much lower precision and recall for the positive class, considerably lower precision for the neutral class, and considerably lower recall for the negative class. 
When looking at the errors made by both models (see the confusion matrices in Figure~\ref{fig:sentiment_confusion}), we saw that while the RoBERTa model tended to confuse the positive or negative posts as neutral, i.e., it tended to overpredict the neutral label, VADER more often confused positive and negative posts. The latter is the limitation of its lexicon-based approach due to its reliance on a static list of sentiment-laden words without considering the broader context in which they appear. 
Because for our study, we considered confusing negative posts as positive and vice versa as a more severe error than predicting neutral instead of either valence, we chose the RoBERTa model despite its somewhat lower accuracy.

\begin{figure*}[ht!]
\centering
\includegraphics[width=1\textwidth]{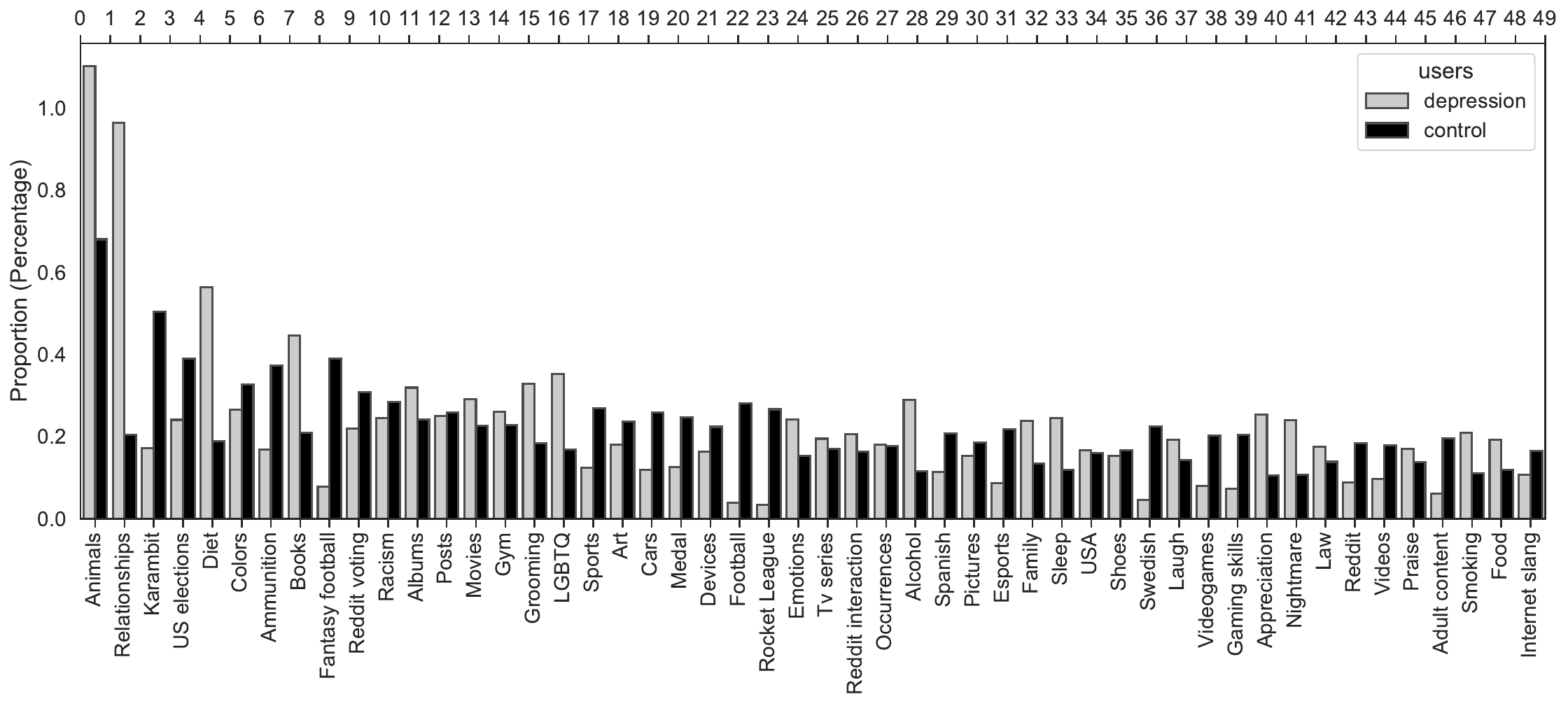}
\caption{Frequency distribution of top 50 topics.}
\label{Fig:topic_distri}
\end{figure*}

\section{Topic modeling}
\label{appendix:Topicmodeling}
We explored the topics by looking 1) at the frequency of posts by user groups and 2) the sentiment distribution of each topic across user groups. 

\begin{figure*}[ht!]
\centering
    \includegraphics[width=1\textwidth]{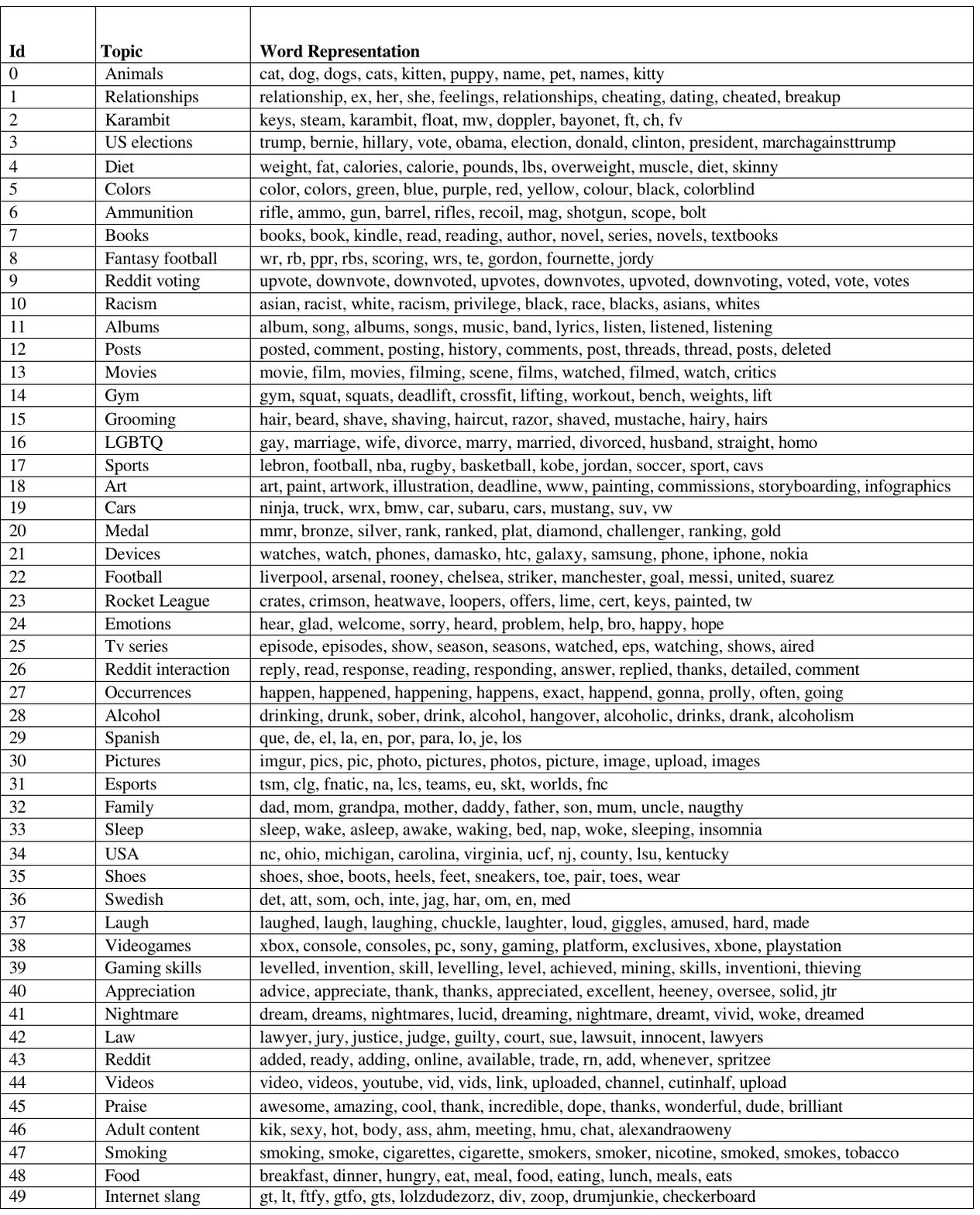}
    \caption{Top 50 topics and their word representations.}
    \label{Fig:fig}
\end{figure*}

\paragraph{Topic frequencies:} The frequency distribution of the top 50 most frequent topics is shown in Figure~\ref{Fig:topic_distri}, which shows the proportion of posts for each topic, normalized by the total posts per group.
These top 50 topics represent 11.3\% of the depression posts and 11.2\% of the control posts. See Figure \ref{Fig:fig} for more details about the topics and their word representations, as extracted from the BERTopic.

As shown in Figure \ref{Fig:topic_distri}, in our dataset, the depression group discusses topics that are related to, for example, animals, relationships, dieting, books, music albums, movies, grooming, LGBTQ themes, and emotions more compared to controls.
Whereas the control group discusses topics related to Karambit/Gaming items, US elections, colors, ammunition, Reddit voting, cars, Online Gaming medals, football, rocket league/Online gaming items trade more compared to depression.

\begin{figure*}[ht!]
    \centering
     \includegraphics[width=1\textwidth]{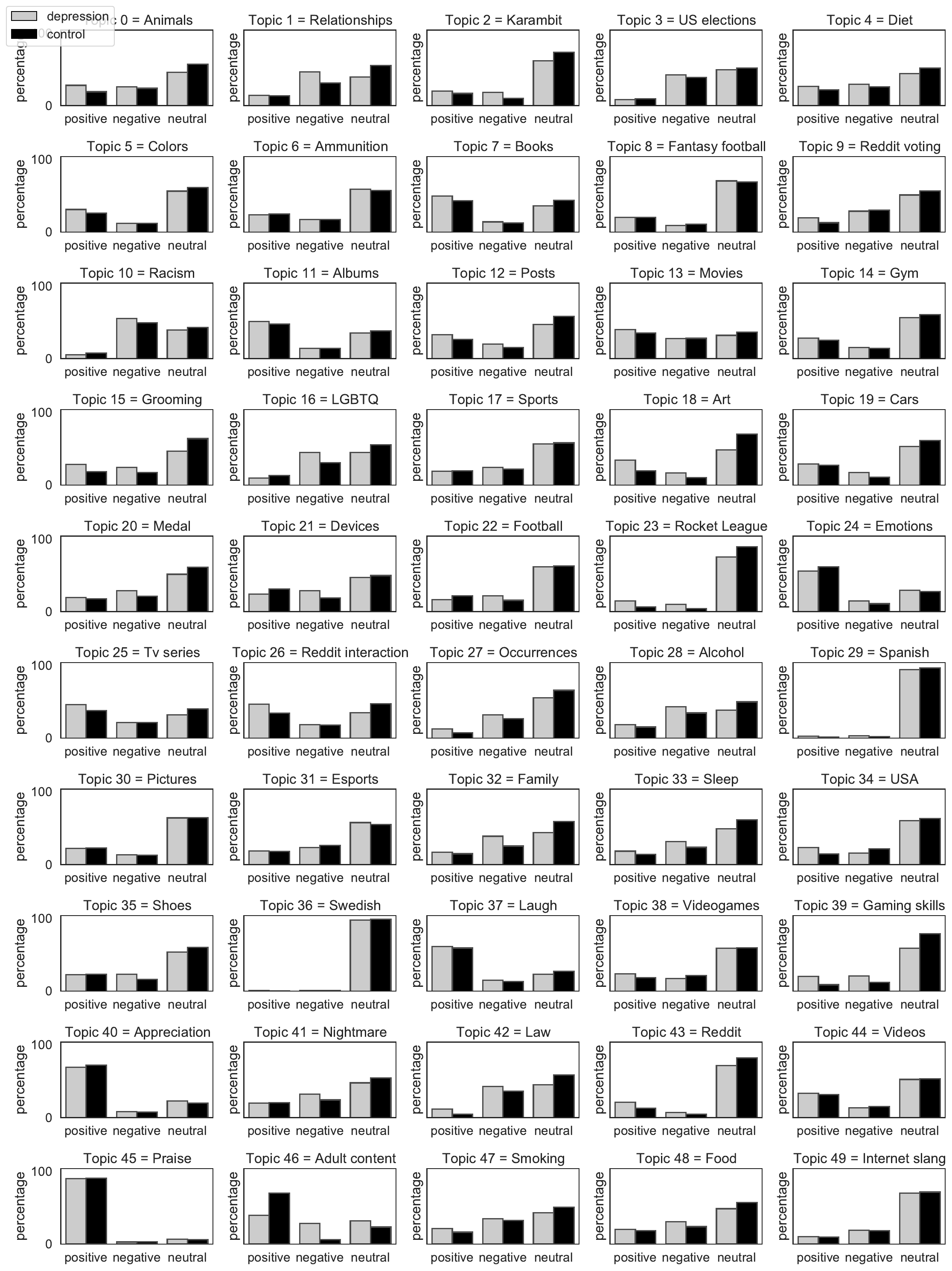}
        \caption{Top 50 topics and their sentiment distributions.}
    \label{Fig:fig_sen}
\end{figure*}

\paragraph{Topic and Sentiment Interactions:} In addition to topics having different frequency distributions by groups, they also show different sentiment patterns. Figure~\ref{Fig:fig_sen} shows the sentiment distribution of both groups for the top 50 topics. 
For instance, regarding discussions related to animals, the depression group expressed more positive sentiments than the control group. On the other hand, for the relationship-related topic, the depression group expressed more negative sentiments. On the topic of US elections, although it is more frequent in the control group, the depression group expresses slightly more negative sentiments.
In topics related to relationships, family, and LGBTQ, the depression and control users express a similar proportion of positive sentiment, while the depression users also express considerably more negative sentiment. In terms of the adult content-related topic, while the most prevalent sentiment is positive for both groups, the control group expresses considerably more positive sentiment and less negative sentiment than the depression group. At the same time, although most sentiments towards the topic related to animals are neutral, the depression group expresses more positive sentiments than the control group. There are several other topics where the depression group expresses the more positive sentiment; the ones with the most visible difference between the groups are related to grooming (topic 15), art (topic 18), TV series (topic 25), Reddit interactions (topic 26) and gaming skills (topic 39).

\section{LIWC}
\label{appendix:LIWC}
Linguistic Inquiry and Word Count (LIWC-22) software, developed by \citet{pennebaker2015development} was used in our research.\footnote{We acknowledge that the use of LIWC is subject to a license, and we have obtained the necessary license for research purposes from the official LIWC website \url{https://www.liwc.app/}}

LIWC contains four summary variables: Analytic Thinking, Clout, Authenticity, and Emotional Tone. The Analytic Thinking variable indicates how people use words that suggest formal, logical, and hierarchical thinking patterns. Clout is an indicator that refers to the relative social status, confidence, or leadership-related language. 
Authenticity shows the degree to which a person is self-monitoring, i.e., spontaneous language use with no self-regulation and filters. Emotional Tone is an indicator of positive and negative emotional tone dimensions. The linguistic dimensions contain attributes representing the percentage of words in a given text containing pronouns, articles, verbs, etc. Psychological processes contain the attributes related to cognitive processes, affect (emotional state, emotional tone), and social processes (social behavior, social references). The extended dictionary contains attributes related to culture, lifestyle, physical health, time orientations, and conversational aspects. A full description of these attributes can be found in \cite{pennebaker2015development}.
The selected 63 attributes and their description are in Table \ref{tab:LIWC}. 

\begin{table*}[t!]
        \small
        \begin{tabular}{p{3.1cm}p{4.3cm}p{2.7cm}p{4.2cm}}
            \toprule
            \bf Category& \bf Description&\bf Category& \bf Description\\
            \cmidrule(lr){1-2}
            \cmidrule(lr){3-4}
             \textit{Summary Variables}&&\textit{Expanded Dictionary}& \\
               \quad Word Count&Total word count&Culture & car, united states, govern*, phone \\
                \quad Analytical thinking&Metric of logical, formal thinking &Lifestyle & work, home, school, working\\
                \quad Clout& Language of leadership, status &\quad Religion& god, hell, christmas*, church\\
                \quad Authentic&Perceived honesty, genuineness &\textit{Physical}\\
                \quad Emotional tone& Degree of positive (negative) tone&\quad \textit{Health}\\
             
             \textit{Linguistic Dimensions}& &\qquad Illness & hospital*, cancer*, sick, pain \\
             \quad pronouns &&\qquad Wellness&  healthy, gym*, supported, diet \\ 
             \qquad 1st person singular& I, me, my, myself & \qquad Mental health& mental health, depressed, suicid*, trauma* \\
             \qquad 1st person plural&we, our, us, lets & \quad Substances & beer*, wine, drunk, cigar* \\
             \qquad 2nd person& you, your, u, yourself& \quad Sexual  &sex, gay, pregnan*, dick \\
             \qquad 3rd person singular& he, she, her, his&\quad  Food  &food*, drink*, eat, dinner* \\
             \qquad 3rd person plural & they, their, them, themsel*&\quad Death  &death*, dead, die, kill \\
             \quad Impersonal pronouns &that, it, this, what&\textit{States}\\
             \quad Auxiliary verbs& is, was, be, have&\quad Need & have to, need, had to, must \\
             \quad Negations& not, no, never, nothing&\quad Want & want, hope, wanted, wish\\

             \textit{Psychological Processes}&&\quad Acquire & get, got, take, getting \\
             Drives & we, our, work, us&\quad Lack & don’t have, didn’t have, *less, hungry \\
             \textit{Cognition}&&\quad Fulfilled & enough, full, complete, extra \\
             \quad All-or-none& all, no, never, always& \quad Fatigue & tired, bored, don’t care, boring \\
             \quad \textit{cognitive processes}&&\textit{Motives}\\
             \qquad Insight& know, how, think, feel&\quad Reward & opportun*, win, gain*, benefit* \\
             \qquad Causation& how, because, make, why&\quad Risk & secur*, protect*, pain, risk* \\
             \qquad Discrepancy& would, can, want, could&\quad  Curiosity & scien*, look* for, research*, wonder \\
             \qquad Tentative& if, or, any, something&\quad Allure & have, like, out, know \\
             \qquad Certitude& really, actually, of course, real&\textit{Perception}&in, out, up, there \\
             \qquad Differentiation& but, not, if, or&quad Feeling & feel, hard, cool, felt \\
             \quad Memory& remember, forget, remind, forgot&\textit{Time orientation}\\
             
             \textit{Affect}&&\quad Past focus& was, had, were, been \\
             \quad \textit{Emotion} &&\quad Present focus &is, are, I’m, can \\
             \qquad Positive emotion& good, love, happy, hope&\quad Future focus& will, going to, have to, may \\
             \qquad Negative emotion& bad, hate, hurt, tired&Conversational & yeah, oh, yes, okay\\
             \qquad \quad Anxiety& worry, fear, afraid, nervous&& \\
             \qquad\quad Anger& hate, mad, angry, frustr*&& \\
             \qquad\quad Sadness &:(, sad, disappoint*, cry&& \\
             \quad Swear words&shit, fuckin*, fuck, damn&& \\
             \textit{Social processes}&&&\\
             \quad Social behavior& said, love, say, care&& \\
             \qquad Politeness&thank, please, thanks, good morning&& \\
             \qquad Interpersonal& conflict fight, kill, killed, attack &&\\
             \qquad Moralization& wrong, honor*, deserv*, judge &&\\
             \qquad Communication& said, say, tell, thank* &&\\
             \quad \textit{Social referents}&&& \\
            \qquad Family& parent*, mother*, father*, baby &&\\
             \qquad Friends&friend*, boyfriend*, girlfriend*, dude&& \\
             \qquad Female references&she, her, girl, woman&&\\
             \qquad Male references& he, his, him, man&&\\
             \bottomrule
             \end{tabular}
        \caption{Selected LIWC attributes and their descriptions.}
    \label{tab:LIWC}
    \end{table*}

\section{Linear regression results}\label{appendix:OLS}
Table \ref{tab:table_ols} represents the impact of various linguistic and psychological attributes on the sentiment difference between the depression and control groups. Attributes associated with positive coefficients indicate a positive influence on sentiment in the depression group. In contrast, the negative coefficients suggest a negative influence on the depression group sentiment.

\begin{table*}[t!]
    \centering
    \begin{tabular}{lSSSSS}
        \toprule
        \bf Attribute & \bf {Coefficient} &\bf {SE} & \bf {t-value} & \bf {p-value} & \bf {95\% CI}\\
        \midrule
        constant & -0.9998 & 0.248 & -4.037& <0.001 &	\numrange{-1.485}{-0.514}\\
        \textit{Summary Variables}&&&& \\
        \quad Word Count & -1.3876 & 0.263 & -5.274 &	<0.001 & \numrange{-1.903}{-0.872} \\
        \quad Clout & 1.1090 & 0.485 & 2.285 & 0.022 &	\numrange{0.158}{2.060} \\
        \quad Emotional tone & 7.2384 & 0.287 & 25.198	 & <0.001 & \numrange{6.675}{7.802} \\
        \midrule
        \textit{Linguistic Dimensions}&&&&  \\
        \quad pronouns&&&&  \\
        \qquad 1st person singular & 0.8189	& 0.377 &	2.175 &	0.030 &	\numrange{0.081}{1.557} \\
        \qquad 3rd person singular & -0.8896 & 0.387 & -2.298 & 0.022 & \numrange{-1.648}{-0.131} \\
        \quad Negations & -0.8863 & 0.304 & -2.916 & 0.004 & \numrange{-1.482}{-0.290} \\
        \midrule
            
        \textit{Psychological Processes}&&&& \\
        Drives & -0.6603 & 0.298 & -2.216 & 0.027 & \numrange{-1.245}{-0.076} \\
        \textit{Cognition}&&&& \\
        \quad \textit{cognitive processes}&&&& \\
        \qquad Causation & -0.9897 & 0.264 & -3.747 & <0.001 & \numrange{-1.508}{-0.472} \\
        \qquad Discrepancy & -0.7955 & 0.320 & -2.485 & 0.013 & \numrange{-1.423}{-0.168} \\
        \qquad Tentative & 0.5659 & 0.271 & 2.086 & 0.037 & \numrange{0.034}{1.098} \\
        \quad Memory & 0.6089 & 0.251 &	2.422 & 0.015 & \numrange{0.116}{1.102} \\
        \textit{Affect}&&&&\\
        \quad \textit{Emotion}&&&& \\
        \qquad Positive emotion & 2.0170 & 0.280 & 7.195 & <0.001 &	\numrange{1.467}{2.567} \\
        \qquad Negative emotion & -1.7006 &	0.420 & -4.045 & <0.001 & \numrange{-2.525}{-0.876} \\
        \qquad\quad Anger & 0.9660 & 0.342 & 2.828 & 0.005 & \numrange{0.296}{1.636} \\
        \qquad\quad Sadness & 0.7130 & 0.296 & 2.408 & 0.016 &\numrange{0.132}{1.294} \\
        \quad Swear words & -2.1978 & 0.272 & -8.086 & <0.001 &	\numrange{-2.731}{-1.665} \\
        \textit{Social processes}&&&&\\
        \quad Social behavior & -0.8867 & 0.416 & -2.134 & 0.033 & \numrange{-1.701}{-0.072} \\
        \qquad Interpersonal & 0.6695 & 0.314 & 2.131 & 0.033 &	\numrange{0.053}{1.286} \\
        \qquad Moralization & -0.6971 & 0.271 & -2.575 & 0.010 & \numrange{-1.228}{-0.166} \\
        \qquad Communication & -0.7260 & 0.348 & -2.086 & 0.037 & \numrange{-1.408}{-0.044} \\
        
        \midrule
        \textit{Expanded Dictionary}&&&&\\    
        \textit{Physical}&&&&\\
        \quad \textit{Health}&&&&\\    
        \quad Substances    & -0.6174 &	0.255 & -2.419 & 0.016 & \numrange{-1.118}{-0.117} \\   
        \quad Death         & -0.6125 &	0.284 &	-2.160 & 0.031 & \numrange{-1.168}{-0.057} \\    
        \textit{Motives}&&&&\\
        \quad Allure        & 0.8486 &	0.278 & 3.049  & 0.002 & \numrange{0.303}{1.394} \\
        \textit{Time orientation}&&&&\\
        \quad Past focus    & -0.5674 &	0.286 &	-1.981 & 0.048 & \numrange{-1.129}{-0.006} \\
        \quad Present focus & -1.0653 &	0.320 &	-3.334 & 0.001 & \numrange{-1.692}{-0.439} \\
             
        \bottomrule
               
        \end{tabular}
        \caption{Linear regression analysis summary.
        }
        \label{tab:table_ols}
\end{table*}

\end{document}